\documentclass{article}

\usepackage[preprint]{neurips_2019}

\usepackage[utf8]{inputenc} 
\usepackage[T1]{fontenc}    
\usepackage{url}            
\usepackage{booktabs}       
\usepackage{amsfonts}       
\usepackage{nicefrac}       
\usepackage{microtype}      

\usepackage{algorithmic}
\usepackage{algorithm}
\usepackage{amsmath}

\usepackage{tikz}
\usetikzlibrary{positioning}
\usetikzlibrary{shapes.geometric,decorations.markings}
\usetikzlibrary{shapes}
\usetikzlibrary{calc}
\usetikzlibrary{fit}
\usetikzlibrary{fadings}
\usetikzlibrary{arrows,decorations.pathreplacing}

\usepackage{algorithm}
\usepackage{algorithmic}
\usepackage{amsmath}
\usepackage{amssymb}

\usepackage{soul}
\usepackage{placeins}
\usepackage{siunitx}

\usepackage{ifthen}
\newboolean{inlr}
\newboolean{midlr}
\setboolean{inlr}{false}

\newcommand\lr[1]{%
  \ifthenelse{\boolean{inlr}}%
    {\ifthenelse{\boolean{midlr}}%
      {\middle#1\setboolean{inlr}{true}}
      {\right#1}}%
    {\left#1\setboolean{inlr}{true}}}%
\newcommand\lrmid{%
  \ifthenelse{\boolean{inlr}}%
    {\setboolean{midlr}{true}}
    {}}%

\usepackage{subfig}

\title{A neurally plausible model learns successor representations in partially observable environments}

\author{%
  Eszter V\'ertes \hspace{0.5cm} Maneesh Sahani \\
  Gatsby Computational Neuroscience Unit\\
  University College London\\
  London, W1T 4JG\\
    \texttt{\{eszter, maneesh\}@gatsby.ucl.ac.uk} \\
}

\begin{document}

\maketitle

\begin{abstract}
Animals need to devise strategies to maximize returns while interacting with their environment based on incoming noisy sensory observations. Task-relevant states, such as the agent's location within an environment or the presence of a predator, are often not directly observable but must be inferred using available sensory information. Successor representations (SR) have been proposed as a middle-ground between model-based and model-free reinforcement learning strategies, allowing for fast value computation and rapid adaptation to changes in the reward function or goal locations.  Indeed, recent studies suggest that features of neural responses are consistent with the SR framework.  However, it is not clear how such representations might be learned and computed in partially observed, noisy environments. Here, we introduce a neurally plausible model using \emph{distributional successor features}, which builds on the distributed distributional code for the representation and computation of uncertainty, and which allows for efficient value function computation in partially observed environments via the successor representation. 
We show that distributional successor features can support reinforcement learning in noisy environments in which direct learning of  successful policies is infeasible. 
%
\end{abstract}

\section{Introduction}

Humans and other animals are able to evaluate long-term consequences of their actions and adapt their behaviour to maximize reward across different environments. This behavioural flexibility is often thought to result from interactions between two adaptive systems implementing model-based and model-free reinforcement learning (RL).

Model-based learning allows for flexible goal-directed behaviour, acquiring an internal model of the environment which is used to evaluate the consequences of actions. As a result, an agent can rapidly adjust its policy to localized changes in the environment or in the reward function. But this flexibility comes at a high computational cost, as optimal actions and value functions depend on expensive simulations in the model.
Model-free methods, on the other hand, learn cached values for states and actions, enabling rapid action selection. This approach, however, is particularly slow to adapt to changes in the task, as adjusting behaviour even to localized changes, e.g. in the placement of the reward, requires updating cached values at all states in the environment.
It has been suggested that the brain makes use both of these complementary approaches, and that they may compete for behavioural control \citep{daw_uncertainty-based_2005}; indeed, several behavioural studies suggest that subjects implement a hybrid of model-free and model-based strategies \citep{daw_model-based_2011, glascher_states_2010}.

Successor representations [SR; \citealp{dayan_improving_1993}] augment the internal state used by model-free systems by the expected future occupancy of each world state. SRs can be viewed as a \emph{precompiled} representation of the model under a given policy. Thus, the SRs fall in between model-free and model-based approaches and can reproduce a range of corresponding behaviours \citep{russek_predictive_2017}.
Recent studies have argued for evidence consistent with SRs in rodent hippocampal and human behavioural data \citep{stachenfeld_hippocampus_2017, momennejad_successor_2017}.

Motivated by both theoretical and experimental work arguing that neural RL systems operate over latent states and need to handle state uncertainty \citep{dayan_decision_2008,gershman_successor_2018,starkweather_dopamine_2017},
our work takes the successor framework further by considering partially observable environments. Adopting the framework of distributed distributional coding \citep{vertes_flexible_2018}, we show how learnt latent dynamical models of the environment can be naturally integrated with SRs defined over the latent space.
We begin with short overviews of reinforcement learning in the partially observed setting (section \ref{sec:pomdp}); the SR (section \ref{sec:SR}); and distributed distributional codes (DDCs) (section \ref{sec:DDC}). In section \ref{sec:DDC-SR}, we describe how using DDCs in the generative and recognition models leads to a particularly simple algorithm for learning latent state dynamics and the associated SR. 

\section{Partially observable Markov decision processes}\label{sec:pomdp}

Markov decision processes (MDP) provide a framework for modelling a wide range of sequential decision-making tasks relevant for reinforcement learning. An MDP is defined by a set of states $S$ and actions $A$, a reward function $R:S\times A\rightarrow\mathbb{R}$, and a probability distribution $\mathcal{T}(s'|s,a)$ that describes the Markovian dynamics of the states conditioned on actions of the agent.  For notational convenience we will take the reward function to be independent of action, depending only on state; but the approach we describe is easily extended to the more general case.
A partially observable Markov decision process (POMDP) is a generalization of an MDP where the Markovian states $s\in S$ are not directly observable to the agent. Instead, the agent receives observations ($o \in O$) that depend on the current \emph{latent} state via an observation process $\mathcal{Z}(o|s)$.
Formally, a POMDP is a tuple: ($S$, $A$, $\mathcal{T}$, $R$, $O$, $\mathcal{Z}$, $\gamma$), comprising the objects defined above and the discount factor $\gamma$.
POMDPs can be defined over either discrete or continuous state spaces.  Here, we focus on the more general continuous case, although the model we present is applicable to discrete state spaces as well.


\section{The successor representation}\label{sec:SR}

As an agent explores an environment, the states it visits are ordered by 
the agent's policy and the transition structure of the world.
State representations that respect this dynamic ordering are likely 
to be more efficient for value estimation and may promote more effective
generalization.
This may not be true of the observed state coordinates.  For instance, a
barrier in a spatial environment might mean that two states with adjacent
physical coordinates are associated with very different values.

\citet{dayan_improving_1993} argued that a natural state space for
model-free value estimation is one where distances between states
reflect the similarity of future paths given the agent's policy.
The successor representation (\citealp{dayan_improving_1993}; SR)
for state $s_{i}$ is defined as the expected discounted sum of future
occupancies for each state $s_{j}$, given the current state $s_{i}$:
\begin{equation}
M^{\pi}(s_{i},s_{j})=\mathbb{E}_{\pi}[\overset{\infty}{\sum_{k=0}}\gamma^{k}\mathbb{I}[s_{t+k}=s_{j}]\mid s_{t}=s_{i}]\,.\label{eq:def_SR}
\end{equation}
That is, in a discrete state space, the SR is a $N\times N$ matrix
where $N$ is the number of states in the environment. The SR depends
on the current policy $\pi$ through the expectation in the right
hand side of eq.~\ref{eq:def_SR}, taken with respect to a (possibly stochastic)
policy $p^{\pi}(a_{t}|s_{t})$ and environment $\mathcal{T}(s_{t+1}|s_{t},a_{t})$.
Importantly, the SR makes it possible to express the value function in a
particularly simple form. Following from eq.~\ref{eq:def_SR} and
the definition of the value function:
\begin{equation}
V^{\pi}(s_{i})=\sum_{j}M^{\pi}(s_{i},s_{j})R(s_{j})\,,\label{eq:value_sr}
\end{equation}
where $R(s_{j})$ is the immediate reward in state $s_{j}$.

The successor matrix $M^{\pi}$ can be learned by TD learning,
in much the same way as TD is used to update value functions.  
In particular, the SR is updated according to a TD error:
\begin{equation}
\delta_{t}(s_{j})=\mathbb{I}[s_{t}=s_{j}]+\gamma{M}^{\pi}(s_{t+1},s_{j})-{M}^{\pi}(s_{t},s_{j})\,,
\end{equation}
which reflects errors in \emph{state predictions} rather than rewards,
a learning signal typically associated
with model-based RL.

As shown in eq.~\ref{eq:value_sr}, the value function can
be factorized into the SR---i.e., information about expected future
states under the policy---and instantaneous reward in each state%
\footnote{Alternatively, for the more general case of action-dependent 
reward, the expected instantaneous reward under the policy-dependent 
action in each state.}.
This modularity enables rapid policy evaluation under changing reward
conditions: for a fixed policy only the reward function needs to be
relearned to evaluate $V^{\pi}(s)$. This contrasts with both
model-free and model-based algorithms, which require extensive experience
or rely on computationally expensive evaluation, respectively, to
recompute the value function.

\subsection{Successor representation using features\label{subsec:SR_with_features}}

The successor representation can be generalized to continuous states
$s\in\mathcal{S}$ by using a set of feature functions $\{\psi_{i}(s)\}$ defined
over $\mathcal{S}$. In this setting, the successor representation
(also referred to as the successor feature representation or SF) encodes expected
feature values instead of occupancies of individual states:
\begin{equation}
M^{\pi}(s_{t},i)=\mathbb{E}[\sum_{k=0}^{\infty}\gamma^{k}\psi_{i}(s_{t+k})\mid s_{t},\pi]\label{eq:def_SF}
\end{equation}
%
%
Assuming that the reward function can be written (or approximated)
as a linear function of the features: $R(s)=w_{rew}^{T}\psi(s)$ (where the 
feature values are collected into a vector $\psi(s)$),
the value function $V(s_{t})$ has a simple form analagous to the discrete case:
\begin{equation}
V^{\pi}(s_{t})=w_{rew}^{T}M^{\pi}(s_{t})\label{eq:value_SF}
\end{equation}

For consistency, we can use linear function approximation with the
same set features as in eq. \ref{eq:def_SF} to parametrize the successor
features $M^{\pi}(s_{t},i)$.
\begin{equation}
M^{\pi}(s_{t},i)\approx\sum_{j}U_{ij}\psi_{j}(s_{t})\label{eq:SF_funapprox}
\end{equation}
The form of the SFs, embodied by the weights $U_{ij}$, can be found by
temporal difference learning:
\begin{align}
\Delta U_{ij} & =\delta_{i}\psi_{j}(s_{t}) & 
\delta_{i} & =\psi_{i}(s_{t})+\gamma M(s_{t+1},i)-M(s_{t},i)
\end{align}

As we have seen in the discrete case, the TD error here signals prediction
errors about features of state, rather than about reward.

\section{Distributed distributional codes}\label{sec:DDC}
Distributed distributional codes (DDC) are a candidate for the neural representation of uncertainty \citep{zemel_probabilistic_1998, sahani_doubly_2003} and recently have been shown to support accurate inference and learning in hierarchical latent variable models \citep{vertes_flexible_2018}. In a DDC, a population of neurons represent distributions in their firing rates implicitly, as a set of expectations:
\begin{align}
    \mu=\mathbb{E}_{p(s)}[\psi(s)]
\end{align}
where $\mu$ is a vector of firing rates, $p(s)$ is the represented distribution, and $\psi(s)$ is a vector of encoding functions specific to each neuron. DDCs can be thought of as representing exponential family distributions with sufficient statistics $\psi(s)$ using their mean parameters $\mathbb{E}_{p(s)}[\psi(s)]$ \citep{wainwright_graphical_2008}.

\section{Distributional successor representation}\label{sec:DDC-SR}

As discussed above, the successor representation can support efficient
value computation by incorporating information about the policy and
the environment into the state representation. However, in more realistic
settings, the states themselves are not directly observable and the
agent is limited to state-dependent noisy sensory information.

In this section, we lay out how the DDC representation for uncertainty
allows for learning and computing with successor representations defined
over latent variables. First, we describe an algorithm for learning
and inference in dynamical latent variable models using DDCs.
We then establish a link between the DDC and successor
features (eq.~\ref{eq:def_SF}) and show how they can be combined
to learn what we call the \emph{distributional successor features}.
We discuss different algorithmic and implementation-related choices
for the proposed scheme and their implications.
\subsection{Learning and inference in a state space model using DDCs}\label{subsec:DDC-SSM}

Here, we consider POMDPs where the state-space transition model is itself
defined by a conditional DDC with means that depend linearly on the preceding
state features. That is, the conditional distribution
describing the latent dynamics implied by 
following the policy $\pi$ can be written in the following form:
\begin{equation}
p^\pi(s_{t+1}|s_{t})\Leftrightarrow\mathbb{E}_{s_{t+1}|s_{t}, \pi}[\psi(s_{t+1})]=T^\pi\psi(s_{t})\label{eq:ddc_ssm}
\end{equation}
where $T^\pi$ is a matrix parametrizing the functional relationship between
$s_{t}$ and the expectation of $\psi(s_{t+1})$ with respect to $p^\pi(s_{t+1}|s_{t})$.


The agent has access only to sensory observations $o_{t}$ at each
time step, and in order to be able to make use of the underlying latent
structure, it has to learn the parameters of generative model $p(s_{t+1}|s_{t})$,
$p(o_{t}|s_{t})$ as well as learn to perform inference in that model.

We consider online inference (filtering), i.e. at each time step $t$
the recognition model produces an estimate $q(s_t|\mathcal{O}_t)$ 
of the posterior distribution
$p(s_{t}|\mathcal{O}_{t})$ given all observations up to time $t$:
$\mathcal{O}_{t}=(o_{1},o_{2},\dots o_{t})$. As in the DDC Helmholtz
machine \citep{vertes_flexible_2018}, these distributions are represented
by a set of expectations---i.e., by a DDC:
\begin{equation}
\mu_{t}(\mathcal{O}_{t})=\mathbb{E}_{q(s_{t}|\mathcal{O}_{t})}[\psi(s_{t})]
\end{equation}
The filtering posterior $\mu_{t}(\mathcal{O}_{t})$ is computed iteratively,
using the posterior in the previous time step $\mu_{t-1}(\mathcal{O}_{t-1})$
and the new observation $o_{t}$. Due to the Markovian structure of
the state space model (see fig. \ref{fig:ddc_ssm}), the recognition
model can be written as a recursive function:
\begin{equation}
\mu_{t}(\mathcal{O}_{t})=f_{W}(\mu_{t-1}(\mathcal{O}_{t-1}),o_{t})\label{eq:ddc_filter}
\end{equation}
with a set of parameters $W$.

\begin{algorithm}[t]
\begin{algorithmic}
\STATE Initialise $T, W$
\WHILE{not converged}
\STATE \textbf{Sleep phase:} 
\STATE sample: $\{s_{t}^{sleep},o_{t}^{sleep}\}_{t=0\dots N}\sim p(\mathcal{S}_N, \mathcal{O}_N)$  
\STATE update $W$: $\Delta W\propto\underset{t}{\sum}(\psi(s_{t}^{sleep})-f_{W}(\mu_{t-1}(\mathcal{O}_{t-1}^{sleep}),o_{t}^{sleep}))\nabla_W f_{W}$ 

\STATE \textbf{Wake phase:} 
\STATE $\mathcal{O}_N \leftarrow$ \{collect observations\} 
\STATE infer posterior $\mu_{t}(\mathcal{O}_{t})=f_{W}(\mu_{t-1}(\mathcal{O}_{t-1}),o_{t})$ 
\STATE update $T$: $\Delta T\propto(\mu_{t+1}(\mathcal{O}_{t+1})-T\mu_{t}(\mathcal{O}_{t}))\mu_{t}(\mathcal{O}_{t})^{T}$
\STATE update observation model parameters
\ENDWHILE
\end{algorithmic}
\caption{Wake-sleep algorithm in the DDC state-space model}\label{alg:wake_sleep}
\end{algorithm}

The recognition and generative
models are updated using an adapted version of the wake-sleep algorithm
\citep{hinton_wake-sleep_1995,vertes_flexible_2018}. In the following,
we describe the two phases of the algorithm in more detail (see Algorithm
\ref{alg:wake_sleep}).

\subsubsection*{Sleep phase}

The aim of the sleep phase is to adjust the parameters of the recognition
model given the current generative model. Specifically, the recognition
model should approximate the expectation of the DDC encoding functions
$\psi(s_{t})$ under the filtering posterior $p(s_{t}|\mathcal{O}_{t}).$
This can be achieved by moment matching, i.e., simulating a sequence
of latent and observed states from the current model and minimizing
the Euclidean distance between the output of the recognition model
and the sufficient statistic vector $\psi(.)$ evaluated at the latent
state from the next time step.
\begin{equation}
W\leftarrow\underset{W}{\textrm{argmin}}\underset{t}{\sum}\|\psi(s_{t}^{sleep})-f_{W}(\mu_{t-1}(\mathcal{O}_{t-1}^{sleep}),o_{t}^{sleep})\|^{2}\label{eq:sleep_update}
\end{equation}
where $\{s_{t}^{sleep},o_{t}^{sleep}\}_{t=0\dots N}\sim p(s_{0})p(o_{0}|s_{0})\underset{t=0}{\overset{N-1}{\prod}}p(s_{t+1}|s_{t},T^{\pi})p(o_{t+1}|s_{t+1})$.

This update rule can be implemented online, and after a sufficiently long
sequence of simulations $\{s_{t}^{sleep},o_{t}^{sleep}\}_{t}$
the recognition model will learn to approximate expectations of the
form: $f_{W}(\mu_{t-1}(\mathcal{O}_{t-1}^{sleep}),o_{t}^{sleep})\approx$ $\mathbb{E}_{p(s_{t}|\mathcal{O}_{t})}[\psi(s_{t})]$,
yielding a DDC representation of the posterior.

\subsubsection*{Wake phase}

In the wake phase, the parameters of the generative model are adapted
such that it captures the sensory observations better. Here, we focus on learning the policy-dependent latent dynamics $p^{\pi}(s_{t+1}|s_{t})$;
the observation model can be learned by the approach of \citep{vertes_flexible_2018}.
Given a sequence of inferred posterior representations $\{\mu_{t}(\mathcal{O}_{t})\}$
computed using wake phase observations, the parameters of the latent
dynamics $T$ can be updated by minimizing a simple predictive cost
function:
\begin{equation}
T\leftarrow\underset{T}{\textrm{argmin}}\underset{t}{\sum}\|\mu_{t+1}(\mathcal{O}_{t+1})-T\mu_{t}(\mathcal{O}_{t})\|^{2}\label{eq:wake_eucl}
\end{equation}
The intuition behind eq.~\ref{eq:wake_eucl} is that for the
optimal generative model the latent dynamics satisfies the following
equality: $T^{*}\mu_{t}(\mathcal{O}_{t})=\mathbb{E}_{p(o_{t+1}|\mathcal{O}_{t})}[\mu_{t+1}(\mathcal{O}_{t+1})]$.
That is, the predictions made by combining the posterior at time $t$
and the prior will agree with the average posterior at the next time
step---making $T^{*}$ a stationary point of the optimization in
eq.~\ref{eq:wake_online}. For further details on the nature of the
approximation implied by the wake phase update and its relationship
to variational learning, see the supplementary material.
In practice, the update can be done online, using gradient steps analogous
to prediction errors:
\begin{equation}
\Delta T\propto(\mu_{t+1}(\mathcal{O}_{t+1})-T\mu_{t}(\mathcal{O}_{t}))\mu_{t}(\mathcal{O}_{t})^{T}\label{eq:wake_online}
\end{equation}
%
%
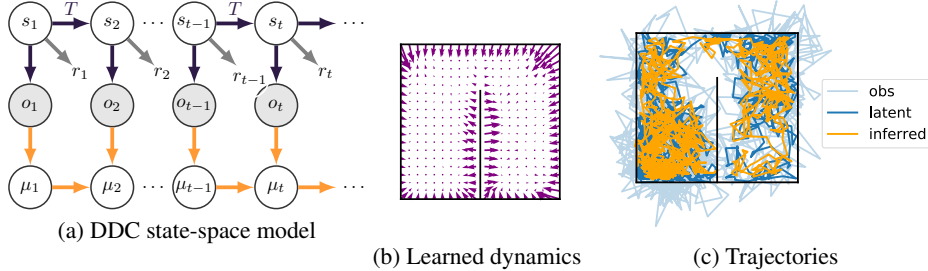
\begin{figure}[th]
\begin{minipage}[c]{0.35\columnwidth}
\subfloat[DDC state-space model]{\centering{}%
\noindent%
\definecolor{colorBlock}{RGB}{46,26,71}
\resizebox{\linewidth}{!}{
\begin{tikzpicture}


\tikzstyle{main}= [circle, minimum size=5ex, inner sep=0pt,draw=black!80, thick]
\tikzstyle{connect}=[line width=0.7mm,color=colorBlock, ->, -latex] 

\tikzstyle{box}=[rectangle, draw=black!100]
\def\nodedist{2em}
\def\nodedisty{2em}

  \node[main] (L1) {$s_1$};
  \node[main] (L2) [right=\nodedist of L1] {$s_2$};
  \node[main] (Lt-1) [right=\nodedist of L2] {$s_{t-1}$};
   \node[main] (Lt) [right=\nodedist of Lt-1] {$s_{t}$};

  \node[main,fill=black!10] (O1) [below=\nodedist of L1] {$o_1$};
  \node[main,fill=black!10] (O2) [right=of O1,below=\nodedist of L2] {$o_2$};
  \node[main,fill=black!10] (Ot-1) [right=of O2,below=\nodedist of Lt-1] {$o_{t-1}$};
  \node[main,fill=black!10] (Ot) [right=of Ot-1,below=\nodedist of Lt] {$o_{t}$};
  \path (L1) edge [connect, edge label=$T$] (L2)
        (L2) -- node[auto=false]{$\ldots$} (Lt-1)
        (Lt-1) edge [connect, edge label=$T$] (Lt);
 
  \path (L1) edge [connect] (O1);
  \path (L2) edge [connect] (O2);
  \path (Lt-1) edge [connect] (Ot-1);
  \path (Lt) edge [connect] (Ot);
  \draw[dashed]  [below=of L1,above=of O1];

  \node (dots) [right=2em of Lt] {$\ldots$};
  \path (Lt) edge [connect] (dots);

  \node[main, draw=white,minimum size=1ex] (r1) [below right=\nodedisty of L1] {$r_1$};
  \node[main, draw=white,minimum size=1ex] (r2) [below right=\nodedisty of L2] {$r_2$};
  \node[main, draw=white,minimum size=1ex] (rt-1) [below right=\nodedisty of Lt-1] {$r_{t-1}$};
  \node[main, draw=white,minimum size=1ex] (rt) [below right=\nodedisty  of Lt] {$r_{t}$};
  
  \path (L1) edge [connect, line width=0.7mm, color=gray] (r1) node {};
  \path (L2) edge [connect, line width=0.7mm, color=gray] (r2);
  \path (Lt-1) edge [connect,line width=0.7mm, color=gray] (rt-1);
  \path (Lt) edge [connect,line width=0.7mm, color=gray] (rt);



  %


 
  \node[main] (L1) [below=\nodedist of O1] {$\mu_1$};
  \node[main] (L2) [below=\nodedist of O2] {$\mu_2$};
  \node[main] (Ltp) [below=\nodedist of Ot-1] {$\mu_{t-1}$};
   \node[main] (Lt) [below=\nodedist of Ot] {$\mu_{t}$};

\tikzstyle{connect}=[line width=0.7mm, color=orange!80, ->, -latex ] 

   \path (L1) edge [connect] (L2)
        (L2) -- node[auto=false]{$\ldots$} (Ltp)
        (Ltp) edge [connect] (Lt);

  \node (dots) [right=2em of Lt] {$\ldots$};
  \path (Lt) edge [connect] (dots);

  \path (O1) edge [connect] (L1);
  \path (O2) edge [connect] (L2);
  \path (Ot-1) edge [connect] (Ltp);
  \path (Ot) edge [connect] (Lt);
  
\end{tikzpicture}
}
}
\end{minipage}
\begin{minipage}[c]{0.6\columnwidth}
\subfloat[Learned dynamics]{\includegraphics[width=3cm]{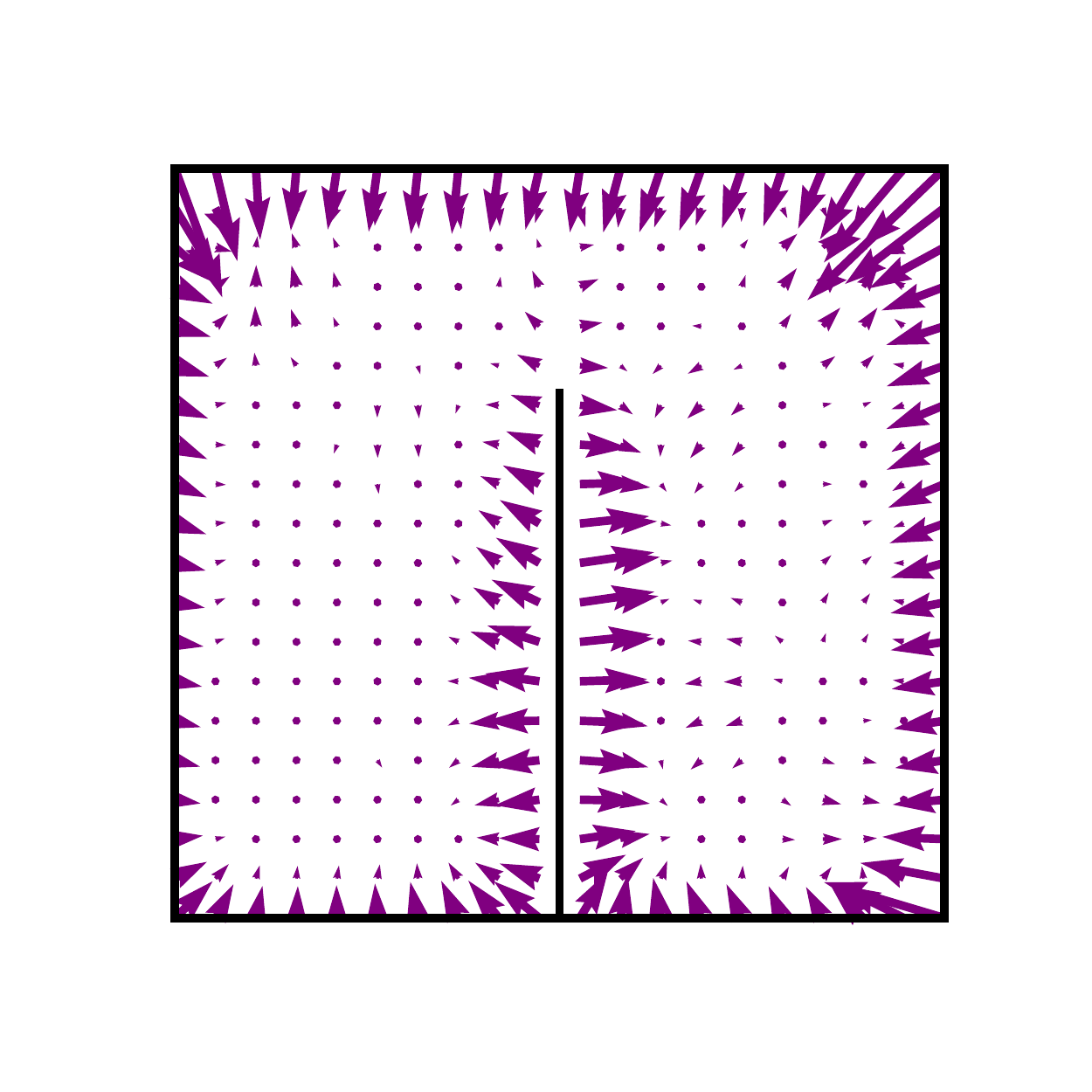}}
%
\subfloat[Trajectories]{
\includegraphics[width=4.5cm]{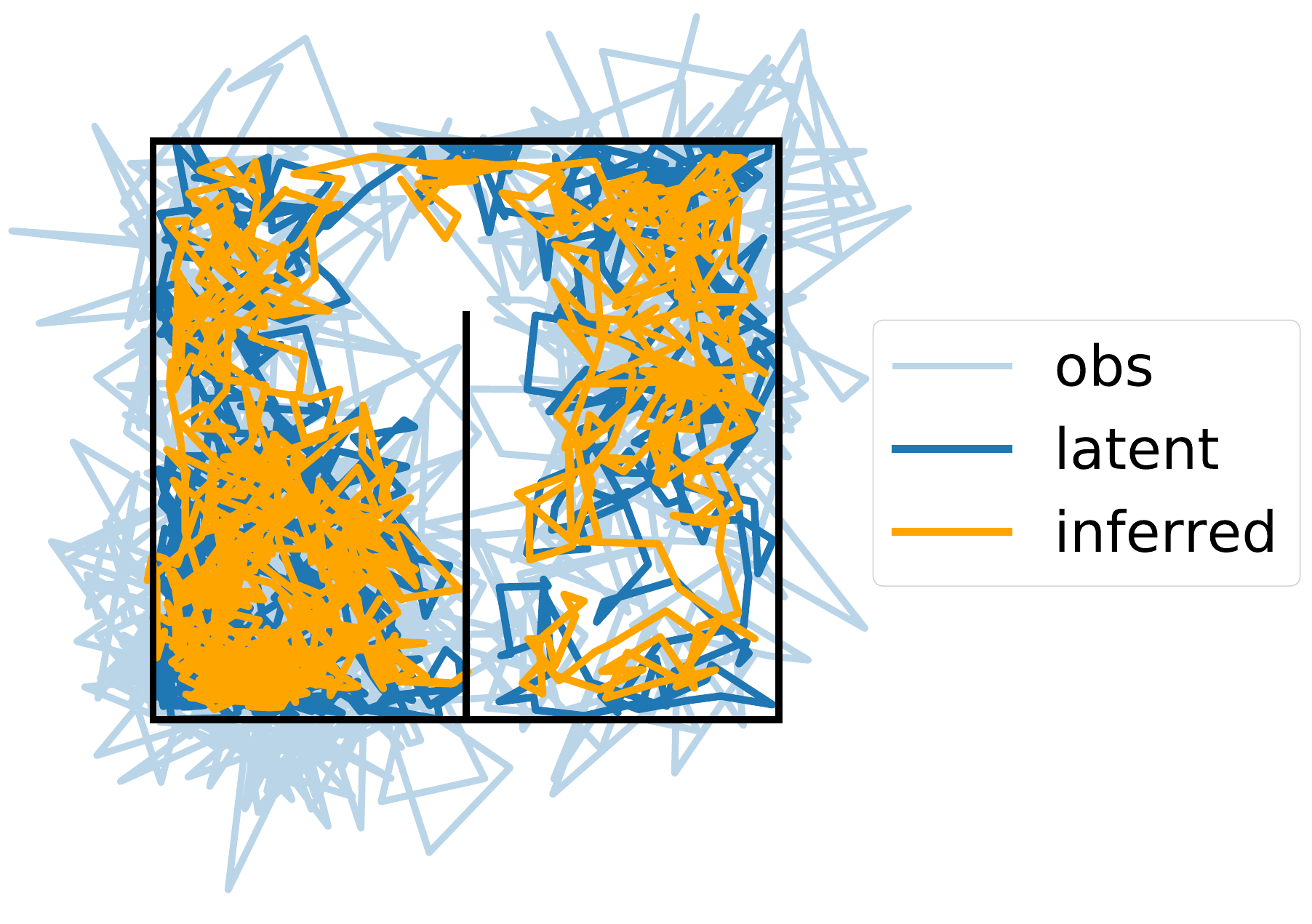}
}
\end{minipage}
\centering{}\caption{Learning and inference in a state-space model parametrized by a DDC. (a) The structure of the generative and recognition models. (b) Visualization of the dynamics $T$ learned by the wake-sleep (algorithm \ref{alg:wake_sleep}). Arrows show the conditional mean $\mathbb{E}_{s_{t+1}|s_t}[s_{t+1}]$ for each location. (c) Posterior mean trajectories inferred using the recognition model, plotted on top of true latent and observed trajectories.}
\label{fig:ddc_ssm}
\end{figure}

Figure \ref{fig:ddc_ssm} shows a state-space model corresponding to a random walk policy in the latent space with noisy observations, learned using DDCs (Algorithm \ref{alg:wake_sleep}). For further details of the experiment, see the supplementary material.
\subsection{Learning distributional successor features}

Next, we show how using a DDC to parametrize the generative model
(eq. \ref{eq:ddc_ssm}) allows for computing the successor features
defined in the latent space in a tractable form, and how this computation
can be combined with inference based on sensory observations.

Following the definition of the SFs (eq.~\ref{eq:def_SF}):
\begin{align}
M(s_{t}) & =\mathbb{E}[\sum_{k=0}^{\infty}\gamma^{k}\psi(s_{t+k})|s_{t},\pi]
=\sum_{k=0}^{\infty}\gamma^{k}\mathbb{E}[\psi(s_{t+k})|s_{t,}\pi] \label{eq:SF_sum}
\end{align}

We can compute the conditional expectations of the feature vector
$\psi$ in eq.~\ref{eq:SF_sum} by applying the dynamics $k$ times to the features
$\psi(s_{t})$: $\mathbb{E}_{s_{t+k}|s_{t}}[\psi(s_{t+k})]=T^{k}\psi(s_{t})$.
Thus, we have:
\begin{align}
M(s_{t}) & =\sum_{k=0}^{\infty}\gamma^{k}T^{k}\psi(s_{t})\\
 & =(I-\gamma T)^{-1}\psi(s_{t})\label{eq:SF_analytic}
\end{align}

Eq.~\ref{eq:SF_analytic} is reminiscent of the result for discrete observed
state spaces $M(s_{i},s_{j})=(I-\gamma P)_{ij}^{-1}$ \citep{dayan_improving_1993},
where P is a matrix containing Markovian transition probabilities
between states. In a continuous state space, however, finding a closed
form solution like eq.~\ref{eq:SF_analytic} is non-trivial, as it
requires evaluating a set of typically intractable integrals. The
solution presented here directly exploits the DDC parametrization
of the generative model and the correspondence between the features
used in the DDC and the SFs.

In this framework, we can not only compute the successor features
in closed form in the latent space, but also evaluate the \emph{distributional
successor features}, the posterior expectation of the SFs given a
sequence of sensory observations:
\begin{align}
\mathbb{E}_{s_{t}|\mathcal{O}_{t}}[M(s_{t})] & =(I-\gamma T)^{-1}\mathbb{E}_{s_{t}|\mathcal{O}_{t}}[\psi(s_{t})]\\
 & =(I-\gamma T)^{-1}\mu_{t}(\mathcal{O}_{t})\label{eq:distSR_analytic}
\end{align}

The results from this section suggest a number of different ways the
distributional successor features $\mathbb{E}_{s_{t}|\mathcal{O}_{t}}[M(s_{t})]$
can be learned or computed.

\subsubsection*{Learning distributional SFs during \emph{sleep phase}}

The matrix $U = (I - \gamma T)^{-1}$ needed to compute distributional SFs in
eq.~\ref{eq:distSR_analytic} can be learned from temporal differences in feature
predictions based on \emph{sleep phase} simulated latent state sequences (section~\ref{subsec:SR_with_features}).


\subsubsection*{Computing distributional SFs by \emph{dynamics}}

Alternatively, eq.~\ref{eq:distSR_analytic} can be implemented
as a fixed point of a linear dynamical system, with recurrent connections
reflecting the model of the latent dynamics:
\begin{align}
\tau\dot{x}_{n} & =-x_{n}+\gamma Tx_{n}+\mu_{t}(\mathcal{O}_{t})\label{eq:SF_dynamics}\\
\Rightarrow x_{\infty} & =(I-\gamma T)^{-1}\mu_{t}(\mathcal{O}_{t})
\end{align}

In this case, there is no need to learn $(I-\gamma T)^{-1}$ explicitly
but it is implicitly computed through dynamics. For this to work,
there is an underlying assumption that the dynamical system in eq.
\ref{eq:SF_dynamics} reaches equilibrium on a timescale faster than
that on which the observations $\mathcal{O}_{t}$ evolve.

Both of these approaches avoid having to compute the matrix inverse
directly and allow for evaluation of policies given by a corresponding dynamics matrix $T^\pi$ offline.

\subsubsection*{Learning distributional SFs during \emph{wake phase}}

Instead of fully relying on the learned latent dynamics to compute
the distributional SFs, we can use posteriors computed by the recognition
model during the wake phase, that is, using observed data. We can
define the distributional SFs directly on the DDC posteriors: $\widetilde{M}(\mathcal{O}_{t})=\mathbb{E}_\pi[\sum_{k}\gamma^{k}\mu_{t+k}(\mathcal{O}_{t+k})|\mu_{t}(\mathcal{O}_{t})]$,
treating the posterior representation $\mu_{t}(\mathcal{O}_{t})$
as a feature space over sequences of observations $\mathcal{O}_{t}=(o_{1}\dots o_{t})$.
Analogously to section \ref{subsec:SR_with_features}, $\widetilde{M}(\mathcal{O}_{t})$
can be acquired by TD learning and assuming linear function approximation:
$\widetilde{M}(\mathcal{O}_{t})\approx U\mu_{t}(\mathcal{O}_{t})$.
The matrix $U$ can be updated online, while executing a given policy
and continuously inferring latent state representations using the
recognition model:
\begin{align}
\Delta U & \propto\delta_{t}\mu_{t}(\mathcal{O}_{t})^{T}\\
\delta_{t} & =\mu_{t}(\mathcal{O}_{t})+\gamma M(\mathcal{O}_{t+1})-M(\mathcal{O}_{t})
\end{align}
It can be shown that
$\widetilde{M}(\mathcal{O}_{t})$, as defined here, is equivalent
to $\mathbb{E}_{s_{t}|\mathcal{O}_{t}}[M(s_{t})]$ if the learned
generative model is optimal--assuming no model mismatch--and the
recognition model correctly infers the corresponding posteriors $\mu_{t}(\mathcal{O}_{t})$ (see supplementary material).
In general, however, exchanging the order of TD learning and inference
leads to different SFs. The advantage of learning the distributional
successor features in the wake phase is that even when the model does
not perfectly capture the data (e.g. due to lack of flexibility or
early on in learning) the learned SFs will reflect the structure in
the observations through the posteriors $\mu_{t}(\mathcal{O}_{t})$.

\subsection{Value computation in a noisy 2D environment}

\begin{figure}
\begin{centering}
\includegraphics[width=0.92\linewidth]{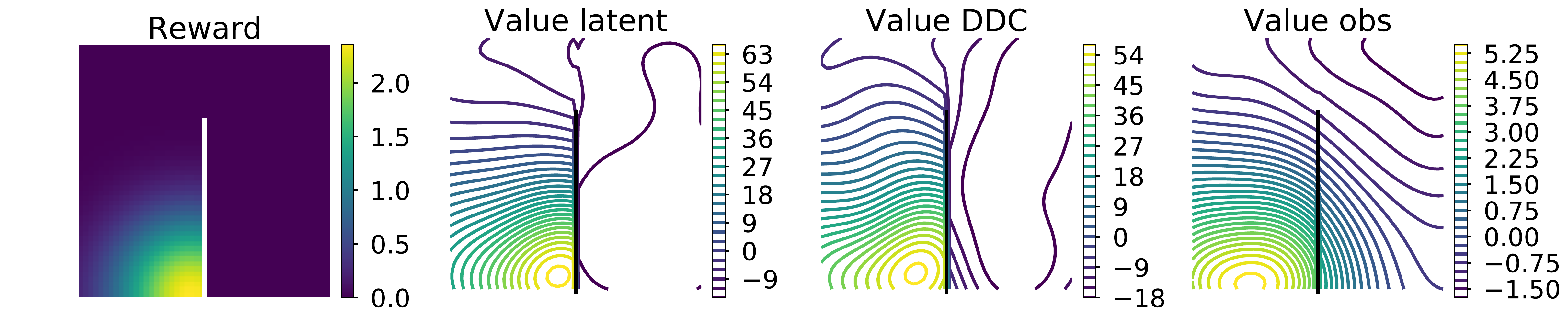}
\includegraphics[width=0.9\linewidth]{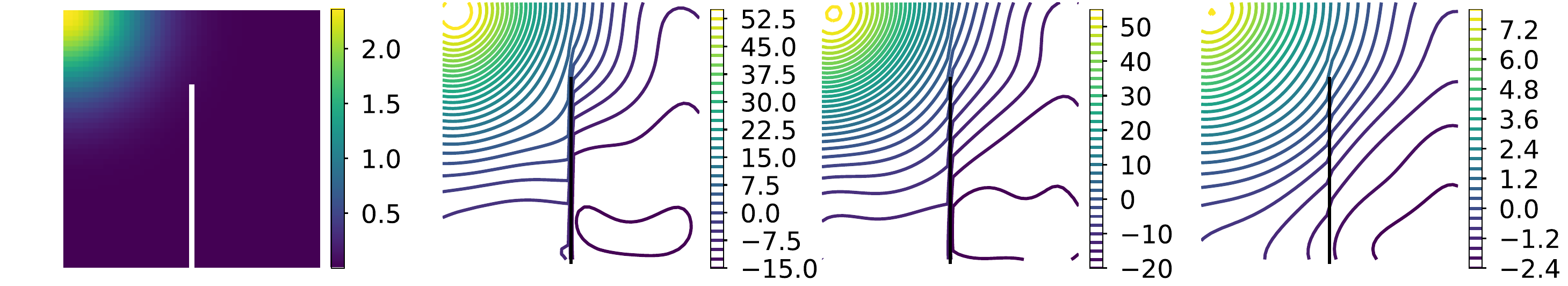}
\caption{Value functions computed using successor features under a random walk policy} 
\label{fig:value_randomwalk}
\par\end{centering}
\end{figure}

We illustrate the importance of being able to consistently handle uncertainty in the SFs by learning value functions in a noisy environment. We use a simple 2-dimensional box environment with continuous state space that includes an internal wall. The agent does not have direct access to its spatial coordinates, but receives observations corrupted by Gaussian noise. 
Figure \ref{fig:value_randomwalk} shows the value functions computed
using the successor features learned in three different settings:
assuming direct access to latent states, treating observations as though 
they were noise-free state measurements, and using latent state estimates
inferred from observations. The value functions computed
in the latent space and computed from DDC posterior representations
both reflect the structure of the environment, while the value function
relying on SFs over the observed states fails to learn about the barrier.

\begin{figure}[th]
\begin{centering}
\includegraphics[width=0.95\linewidth]{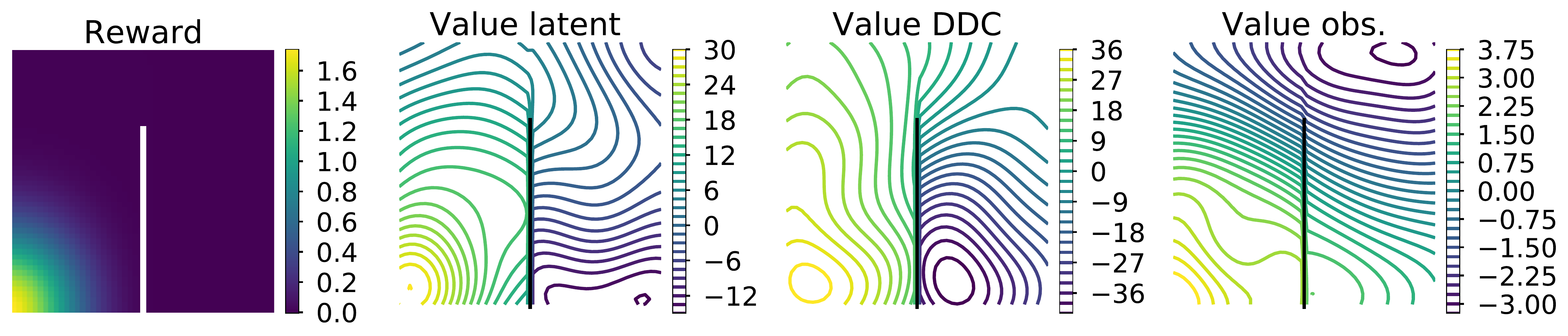}
\par\end{centering}
\begin{centering}
\hspace{-1cm}\includegraphics[width=0.25\linewidth]{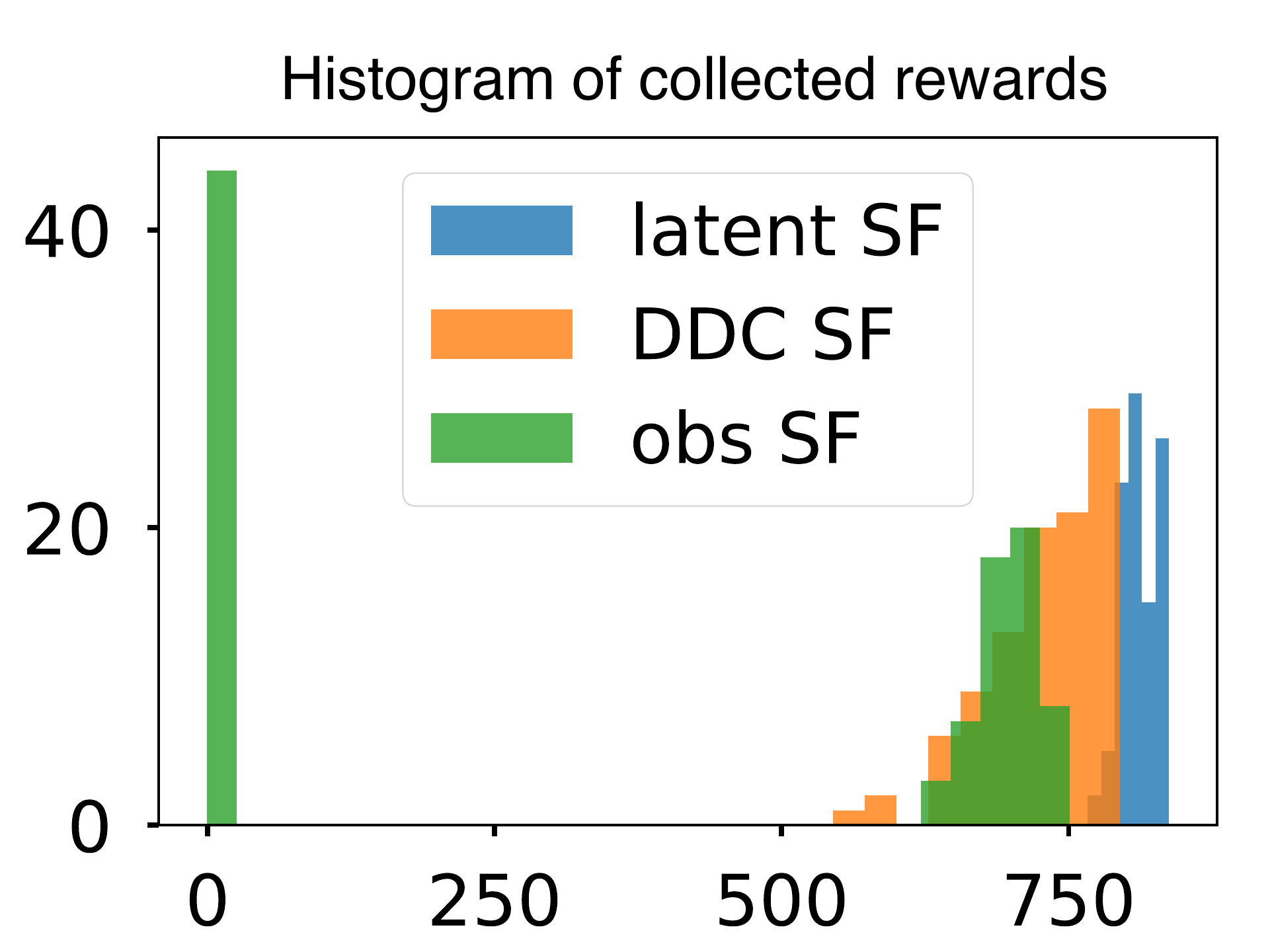}\includegraphics[width=0.2\linewidth]{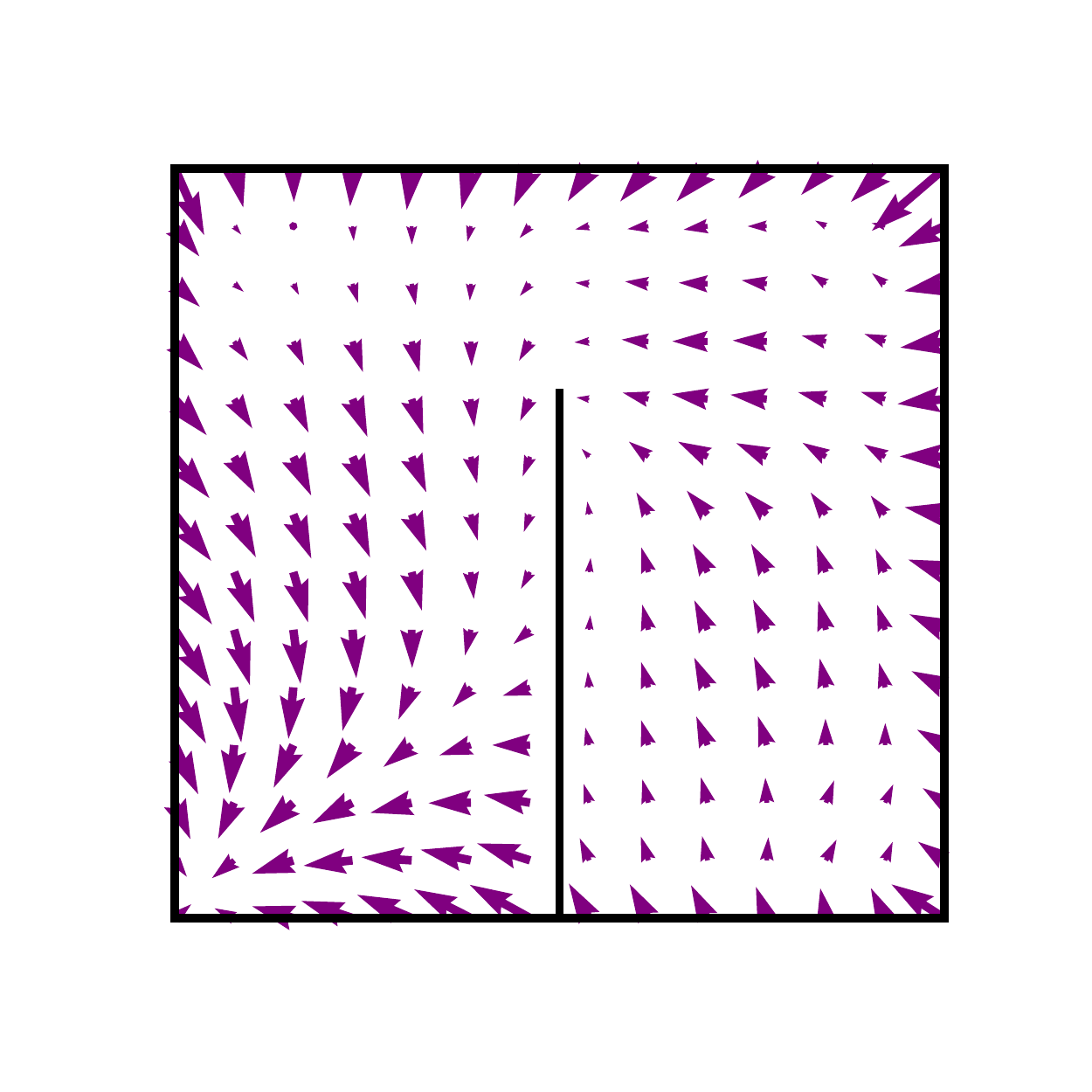}\hspace{0.5cm}\includegraphics[width=0.2\linewidth]{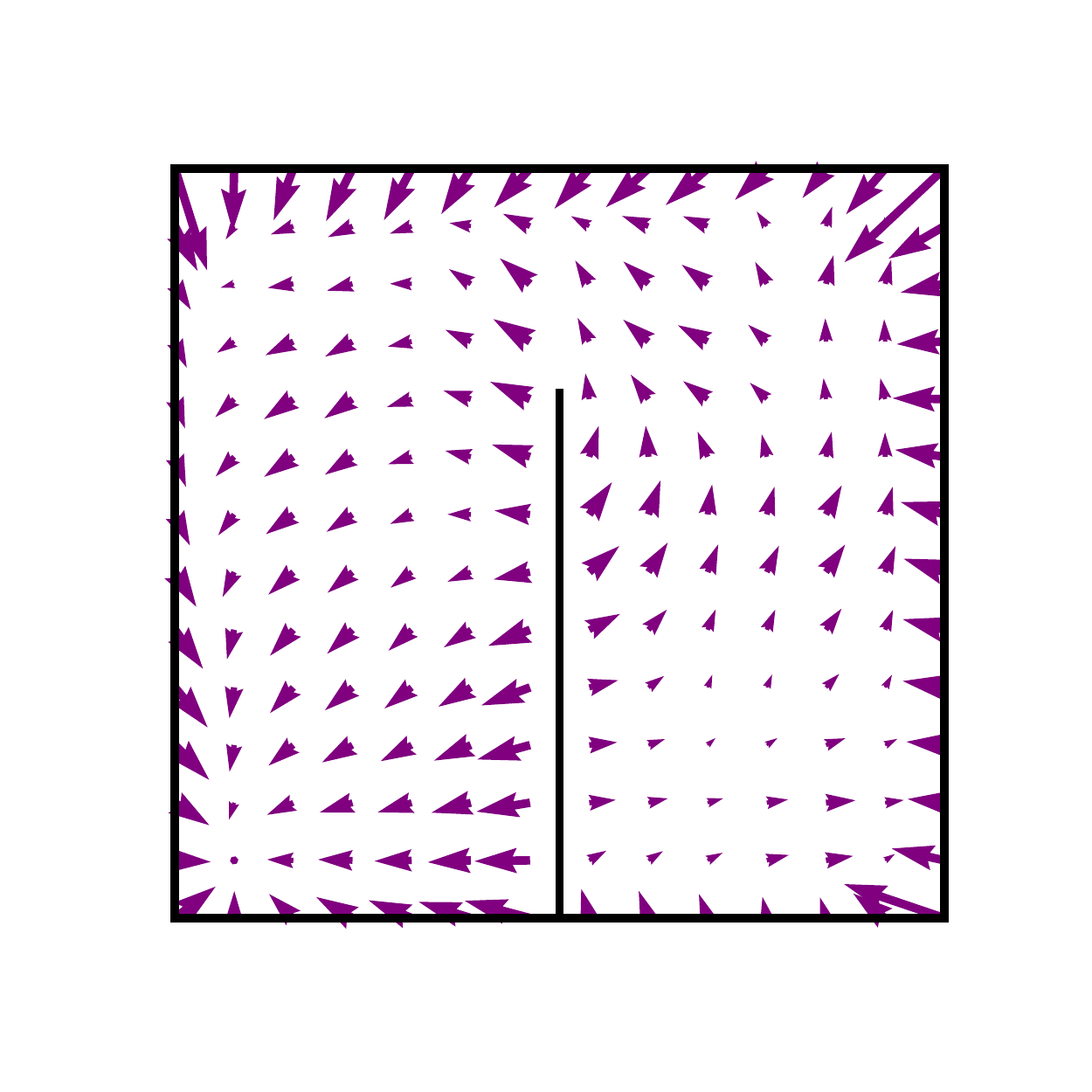}\hspace{0.5cm}\includegraphics[width=0.2\linewidth]{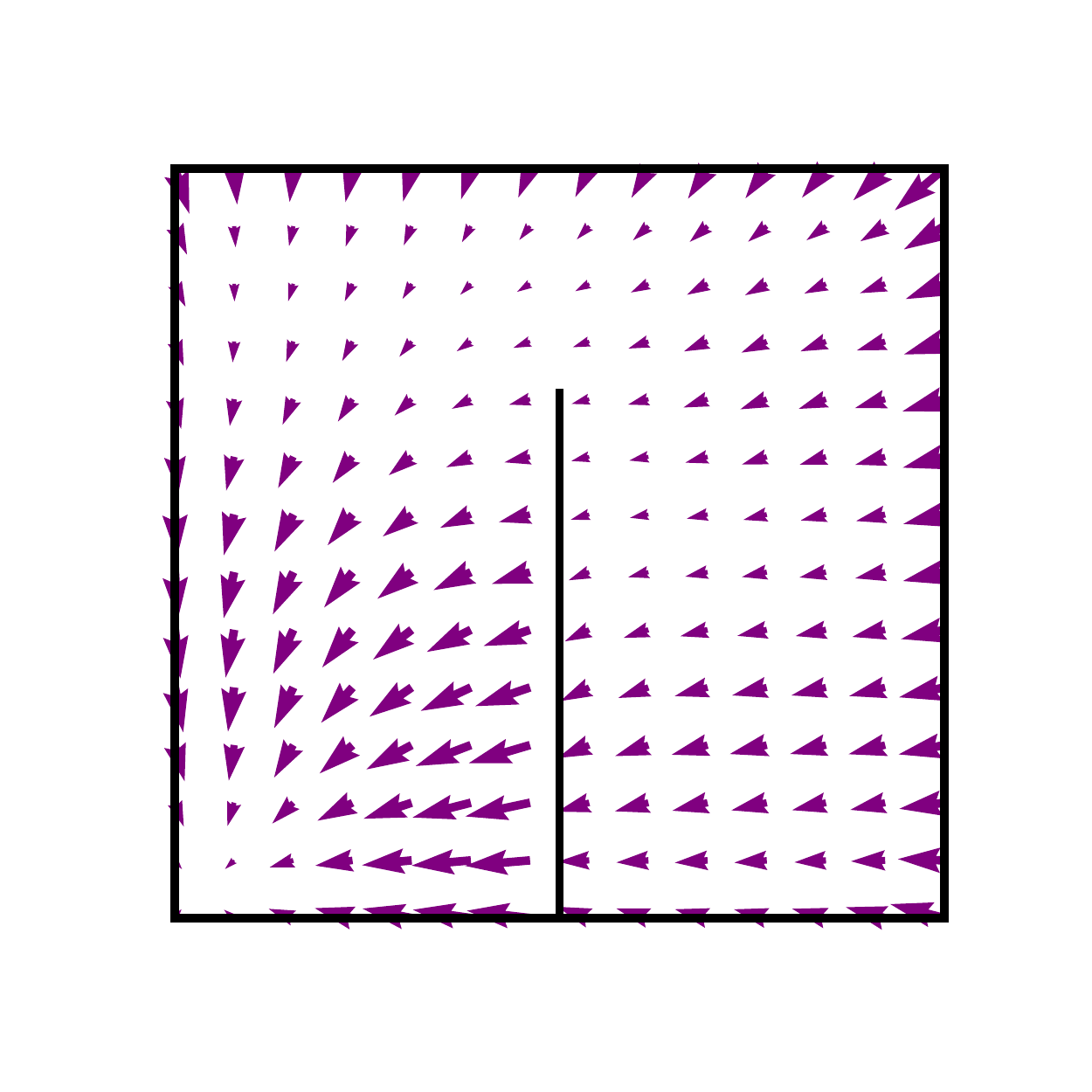}\caption{Value functions computed by SFs under the learned policy. Top row shows reward and value functions learned in the three different conditions. Bottom row shows histogram of collected rewards from 100 episodes with random initial states, and the learned dynamics $T^\pi$ visualized as in  fig. \ref{fig:ddc_ssm}.}
\label{fig:value_policy}
\par\end{centering}
\end{figure}

To demonstrate that this is not simply due to using the suboptimal random
walk policy, but persists through learning, we have learned successor
features while adjusting the policy to a given reward function (see figure \ref{fig:value_policy}). The
policy was learned by generalized policy iteration \citep{sutton_introduction_1998},
alternating between taking actions following a greedy policy and updating
the successor features to estimate the corresponding value function.

The value of each state and action was computed from the value
function $V(s)$ by a one-step look-ahead, combining the immediate reward with the expected
value function having taken a given action:
\begin{equation}
Q(s_t,a_t)=r(s_t)+\gamma\mathbb{E}_{s_{t+1}|s_t,a_t}[V(s_{t+1})]\label{eq:Qsa}
\end{equation}
In our case, as the value function in the latent space is expressed
as a linear function of the features $\psi(s)$: $V(s)=w^{T}U\psi(s)$
(eq. \ref{eq:value_SF}-\ref{eq:SF_funapprox}), the expectation in
\ref{eq:Qsa} can be expressed as:
\begin{align}
\mathbb{E}_{s_{t+1}|s_t,a_t}[V(s_{t+1})] & =w_{\mathrm{rew}}^{T}U\cdot\mathbb{E}_{s'|s,a}[\psi(s_{t+1})]\\
 & =w_{\mathrm{rew}}^{T}U\cdot P\cdot(\psi(s_t)\otimes\phi(a_t))
\end{align}
Where $P$ is a linear mapping, $P:\Psi\times\Phi\rightarrow\Psi$,
that contains information about the distribution $p(s_{t+1}|s_t,a_t)$. More
specifically, $P$ is trained to predict $\mathbb{E}_{s_{t+1}|s_t,a_t}[\psi(s_{t+1})]$
as a bilinear function of state and action features $(\psi(s_t)$, $\phi(a_t)$).
Given the state-action value, we can implement a greedy policy by
choosing actions that maximize $Q(s,a)$:
\begin{align}
a^{*} & =\underset{a \in \mathcal{A}}{\textrm{argmax}}Q(s_t,a_t)\\
 & =\underset{a\in \mathcal{A}}{\textrm{argmax }}r(s_t)+ \gamma w_{\mathrm{rew}}^{T}U\cdot P\cdot(\psi(s_t)\times\phi(a_t))\label{eq:argmax}
\end{align}
The argmax operation in eq.~\ref{eq:argmax} (possibly over a continuous
space of actions) could be biologically implemented by a ring attractor where the
neurons receive state-dependent input through feedforward
weights reflecting the tuning ($\phi(a)$) of each neuron in the ring.

Just as in figure \ref{fig:value_randomwalk}, we compute the value function in the fully observed case, using inferred states or using only the noisy observations. For the latter two, we replace $\psi(s_t)$ in eq.~\ref{eq:argmax}
with the inferred state representation $\mu(\mathcal{O}_t)$ and the observed features $\psi(o_t)$, respectively. As the agent follows the greedy policy and it receives new observations the corresponding SFs are adapted accordingly.
Figure \ref{fig:value_policy} shows the learned value functions $V^\pi(s)$, $V^\pi(\mu)$ and $V^\pi(o)$ for a given reward location and the corresponding dynamics $T^\pi$. The agent having access to the true latent state as well as the one using distributional SFs successfully learn policies leading to the rewarded location. As before, the agent learning SFs purely based on observations remains highly sub-optimal.

\section{Discussion}



We have shown that representing uncertainty over latent variables using DDCs can be naturally integrated with representations of uncertainty about future states and therefore can generalize SRs to more realistic environments with partial observability.

In our work, we have defined distributional SFs over states, using single step look-ahead to compute state-action values (eq. \ref{eq:Qsa}). Alternatively, SFs could be defined directly over both states and actions \citep{kulkarni_deep_2016,barreto_successor_2017} with the distributional development presented here.
\cite{barreto_successor_2017, barreto_transfer_2019} has shown that successor representations corresponding to previously learned tasks can be used as a basis to construct policies for novel tasks, enabling generalization. Our framework can be extended in a similar way, eliminating the need to adapt the SFs as the policy of the agent changes.



The framework for learning distributional successor features presented here makes a number of connections to experimental observations in the hippocampal literature.  While it has been argued that the hippocampus holds an internal model of the environment and thereby supports model-based decision making \citep{miller_dorsal_2017}, there is little known about how such a model is acquired.
Hippocampal replays observed in rodents during periods of immobility and sleep have been interpreted as mental simulations from an internal model of the environment, and therefore a neural substrate for model-based planning \citep{pfeiffer_hippocampal_2013, mattar_prioritized_2018}.
Here, we propose a complementary function of replays that is to do with learning in the context of partially observed environments. The replayed sequences could serve to refine the recognition model to accurately infer distributions over latent states, just as in the \emph{sleep phase} of our algorithm. Broadly consistent with this idea, \citet{stella_hippocampal_2019} recently observed replays reminiscent of random walk trajectories after an animal freely explored the environment. These paths were not previously experienced by the animal, and could indeed serve as a training signal for the recognition model. Learning to perform inference is itself a prerequisite for learning the dynamics of the latent task-relevant variables, i.e. the internal model.











\bibliographystyle{plainnat}
\bibliography{bibl_west}

 \newpage

\begin{center}
    \LARGE\textbf{Supplementary material}
\end{center}

\appendix
\section{Approximations in the wake phase update\label{subsec:appendix_wake}}

Here, we give some additional insights into the nature of the approximation
implied by the wake phase update for the DDC state-space model and
discuss its link to variational methods.

According to the standard M step in variational EM, the model parameters
are updated to maximize the expected log-joint of the model under
the approximate posterior distributions:
\begin{align}
\Delta\theta & \propto\nabla_{\theta}\sum_{t}\mathbb{E}_{q(s_{t},s_{t+1}|\mathcal{O}_{t+1})}[\log p_{\theta}(s_{t+1}|s_{t})]\\
 & =\nabla_{\theta}\sum_{t}-\int q(s_{t},s_{t+1}|\mathcal{O}_{t+1})(\log p_{\theta}(s_{t+1}|s_{t})+\log q(s_{t}|\mathcal{O}_{t+1}))d(s_{t},s_{t+1})\\
 & =\nabla_{\theta}\sum_{t}-KL[q(s_{t},s_{t+1}|\mathcal{O}_{t+1})\|p_{\theta}(s_{t+1}|s_{t})q(s_{t}|\mathcal{O}_{t+1})]\label{eq:Mstep_KL}
\end{align}

After projecting the distributions appearing in the KL divergence
(eq. \ref{eq:Mstep_KL}) into the joint exponential family defined
by sufficient statistics $[\psi(s_{t}),\psi(s_{t+1})]$, they can
be represented using the corresponding mean parameters:

\begin{align}
q(s_{t},s_{t+1}|\mathcal{O}_{t+1}) & \stackrel{\mathcal{P}}{\Longrightarrow}\left[\begin{array}{c}
\mathbb{E}_{q(s_{t},s_{t+1}|\mathcal{O}_{t+1})}[\psi(s_{t})]\\
\mathbb{E}_{q(s_{t},s_{t+1}|\mathcal{O}_{t+1})}[\psi(s_{t+1})]
\end{array}\right]=\left[\begin{array}{c}
\mu_{t}(\mathcal{O}_{t+1})\\
\mu_{t+1}(\mathcal{O}_{t+1})
\end{array}\right]\label{eq:expfam_proj1}\\
p_{\theta}(s_{t+1}|s_{t})q(s_{t}|\mathcal{O}_{t+1}) & \stackrel{\mathcal{P}}{\Longrightarrow}\left[\begin{array}{c}
\mathbb{E}_{p_{\theta}(s_{t+1}|s_{t})q(s_{t}|\mathcal{O}_{t+1})}[\psi(s_{t})]\\
\mathbb{E}_{p_{\theta}(s_{t+1}|s_{t})q(s_{t}|\mathcal{O}_{t+1})}[\psi(s_{t+1})]
\end{array}\right]=\left[\begin{array}{c}
\mu_{t}(\mathcal{O}_{t+1})\\
T\mu_{t}(\mathcal{O}_{t+1})
\end{array}\right]\label{eq:expfam_proj2}
\end{align}

To restrict ourselves to online inference, we can make a further approximation:
$\mu_{t}(\mathcal{O}_{t+1})\approx\mu_{t}(\mathcal{O}_{t})$. Thus,
the wake phase update can be thought of as replacing the KL divergence
in equation \ref{eq:Mstep_KL} by the Euclidean distance between the
(projected) mean parameter representations in eq. \ref{eq:expfam_proj1}-\ref{eq:expfam_proj2}.

\begin{equation}
\sum_{t}\|\mu_{t+1}(\mathcal{O}_{t+1})-T\mu_{t}(\mathcal{O}_{t})\|^{2}
\end{equation}

Note that this cost function is directly related to the maximum mean
discrepancy (\cite{gretton_kernel_2012}; MMD)--a non-parametric
distance metric between two distributions--with a finite dimensional
RKHS.

\section{Equivalence of $\mathbb{E}_{p(s_{t}|\mathcal{O}_{t})}[M(s_{t})]$
and $\widetilde{M}(\mu_{t}(\mathcal{O}_{t}))$\label{subsec:Equivalence-of-M}}

\begin{align}     
\widetilde{M}(\mu_t(\mathcal{O}_t))  &= \mathbb{E}_{p(\mathcal{O}_{>t}|\mathcal{O}_t)}[\sum_k \gamma^k  \mu_{t+k}(\mathcal{O}_{t+k})]\\
\textrm{where }\nonumber \\     
\mathbb{E}_{p(\mathcal{O}_{>t}|\mathcal{O}_t)} [\mu_{t+k}(\mathcal{O}_{t+k})] &=  \mathbb{E}_{p(\mathcal{O}_{t+1:t+k}|\mathcal{O}_t)} [\mu_{t+k}(\mathcal{O}_{t+k})] \label{eq:Emu}\\     
&= \int d \mathcal{O}_{t+1:t+k} \textrm{ } p(\mathcal{O}_{t+1:t+k}|\mathcal{O}_t) \int d s_{t+k}\textrm{ } p(s_{t+k}|\mathcal{O}_{t+k}) \psi(s_{t+k}) \\     
&= \int d \mathcal{O}_{t+1:t+k} \textrm{ } p(\mathcal{O}_{t+1:t+k}|\mathcal{O}_t) \int d s_{t+k}\textrm{ } \frac{p(s_{t+k}, \mathcal{O}_{t+1:t+k}|\mathcal{O}_{t})}{p(\mathcal{O}_{t+1:t+k}|\mathcal{O}_t)} \psi(s_{t+k})\\     
&= \int d s_{t+k}\int d \mathcal{O}_{t+1:t+k}\textrm{ } p(s_{t+k}, \mathcal{O}_{t+1:t+k}|\mathcal{O}_{t}) \psi(s_{t+k})\\     
&= T^k \mu_t \label{eq:Tk_mu} \\ 
\end{align}
Thus we have:
\begin{align}
\widetilde{M}(\mu_t(\mathcal{O}_t))  &= \sum_k \gamma^k T^k \mu_t \\
&= (I - \gamma T)^{-1} \mu_t \\
&= \mathbb{E}_{p(s_t|\mathcal{O}_t)}[M(s_t)]
\end{align}

\section{Further experimental details}
\textbf{Figure 1: Learning and inference in the DDC state-space model}

The generative model corresponding to a random walk policy:
\begin{align}
p(s_{t+1}|s_t) &= [s_t + \Tilde{\eta} ]_{\textsc{walls}}, \textrm{ } \\
p(o_t|s_t) &= s_t + \xi
\end{align}
Where $[.]_{\textsc{walls}}$ indicates the constraints introduced by the walls in the environment (outer walls are of unit length).  ${\eta} \sim \mathcal{N}(0,\sigma_s=1.), \Tilde{\eta}=0.06*\eta/\|\eta\|$, $\xi \sim \mathcal{N}(0,\sigma_o=0.1)$, $s_t,o_t \in \mathbb{R}^2$

We used K=100 Gaussian features with width $\sigma_\psi=0.3$ for both the latent and observed states. A small subset of features were truncated along the internal wall, to limit the artifacts from the function approximation. Alternatively, a features with various spatial scales can also be used. The recursive recognition model was parametrized linearly using the features:
\begin{align}
f_W(\mu_{t-1}, o_t) = W [T\mu_{t-1}; \psi(o_t)]
\end{align}
As sampling from the DDC parametrized latent dynamics is not tractable in general, in the sleep phase, we generated approximate samples from a Gaussian distribution with consistent mean. 
The generative and recognition models were trained through 50 wake-sleep cycles, with $3\cdot10^4$ sleep samples, and $5\cdot10^4$ wake phase observations.

The latent dynamics in Fig.1b is visualized by approximating the mean dynamics as a linear readout from the DDC: $\mathbb{E}_{s_{t+1}|s_t}[s_{t+1}]\approx \alpha T\psi(s_t)$ where $s\approx \alpha\psi(s)$.

\paragraph{Figure 2}
To compute the value functions under the random walk policy we computed the SFs based on the latent ($\psi(s)$), inferred ($\mu$) or observed ( $\psi(o)$) features, with discount factor $\gamma=0.99$. In each case, we estimated the reward vector $w_{\mathrm{rew}}$ using the available state information.

\paragraph{Figure 3}
To construct the state-action value function, we  used 10 features over actions $\phi(a)$, von Mises functions ($\kappa=2.$) arranged evenly on $[0,2\pi]$.
The policy iteration was run for 500 cycles, and in each cycle an episode of 500 steps was collected according to the greedy policy. The visited latent, inferred of observed state sequences were used to update the corresponding SFs to re-evaluate the policy. To facilitate faster learning, only episodes with positive returns were used to update the SFs.

\end{document}